\documentclass[runningheads]{llncs}
\usepackage{graphicx}

\usepackage[english]{babel}
\usepackage{hyperref}
\addto\extrasenglish{
}

\usepackage{wrapfig,amssymb,amsmath}
\usepackage{cite}
\usepackage{booktabs}

\DeclareMathOperator*{\argmax}{\arg\!\max}

\begin{document}
\title{DartsReNet: Exploring new RNN cells in ReNet architectures}

\author{Brian B. Moser\textsuperscript{1,2}\and
Federico Raue\textsuperscript{1}\and
J{\"o}rn Hees\textsuperscript{1}\and
Andreas Dengel\textsuperscript{1,2}}

\authorrunning{Moser et al.}
\institute{\textsuperscript{1} German Research Center for Artificial Intelligence (DFKI), Germany \\
\textsuperscript{2} TU Kaiserslautern, Germany\\
\email{\{brian.moser, federico.raue, joern.hees, andreas.dengel\}@dfki.de} }
\maketitle              
\begin{abstract}
We present new Recurrent Neural Network (RNN) cells for image classification using a Neural Architecture Search (NAS) approach called DARTS.
We are interested in the ReNet architecture, which is a RNN based approach presented as an alternative for convolutional and pooling steps.
ReNet can be defined using any standard RNN cells, such as LSTM and GRU.
One limitation is that standard RNN cells were designed for one dimensional sequential data and not for two dimensions like it is the case for image classification.
We overcome this limitation by using DARTS to find new cell designs.
We compare our results with ReNet that uses GRU and LSTM cells.
Our found cells outperform the standard RNN cells on CIFAR-10 and SVHN.
The improvements on SVHN indicate generalizability, as we derived the RNN cell designs from CIFAR-10 without performing a new cell search for SVHN.\footnote{The source code of our approach and experiments is available at
\url{https://github.com/Brian-Moser/DartsReNet}.}

\end{abstract}

\section{Introduction}
Convolutional Neural Networks (CNNs) achieved state-of-the-art results on image classification, e.g., GoogLeNet, VGG, and ResNet \cite{szegedy2015going, simonyan2014very, he2016deep}. The current trend of finding new architectures with better performances relies mainly on the basis of convolutional operations.
Nevertheless, an alternative way is using Recurrent Neural Networks (RNNs) as a core element or in combination with CNNs.
They have shown promising results for Computer Vision (CV) tasks, e.g., ReNet\cite{visin2015renet}, MD-LSTM \cite{graves2009offline}, PyraMiD-LSTM  \cite{stollenga2015parallel} or Optical Character Recognition related approaches \cite{breuel2013high}.
RNN approaches have the advantage that they capture of capturing the global context. Nonetheless, RNNs are slower than CNNs because of their sequential nature, which is less parallelizable.

In general, finding suitable Neural Network architectures is an expensive human effort of trial and error until reaching a desirable performance (e.g. number of layers or the size of each layer).
Nowadays, a new field has emerged called Neural Architecture Search (NAS) \cite{zoph2016neural}, which is trying to automate the manual process of designing such architectures.

Recent works of manually designed state-of-the-art architectures showed that repetitions of fixed structures are favorable, e.g., Residual Blocks in ResNet\cite{he2016deep} or Inception-Modules in GoogLeNet \cite{szegedy2015going}.
These structures are smaller-sized graphs and can be stacked to form a larger-scale network architecture.
Across the NAS literature, they are called cells, and a cell-based approach tries to find them. For this work, we have a particular interest in DARTS, which is a cell-based NAS approach \cite{liu2018darts}.
Moreover, DARTS is a differentiable architecture search, so it uses gradient descent to find such cells.

Although NAS approaches show mainly new architectures based on CNNs for CV tasks, there is only work on finding RNN cells for one dimensional sequential data, such as text prediction or speech recognition. We also observe that alternative RNN-based approaches for CV tasks are mostly considering standard RNN cells designs like LSTM or GRU \cite{hochreiter1997long, cho2014learning}. Despite their effectiveness, it is unclear if they are optimal for an image-based domain since the original motivation of LSTM and GRU address one dimensional sequences.

In this work, we are interested in modifying the RNN cell design (which is the core element) in the ReNet model using DARTS.
We further evaluated two more variants for search space: One is \textit{Sigmoid} Weighting on the input sequence to weight the importance of a time step, and the other is Directional Weight Sharing, which uses the same weight matrix for both directions in a bidirectional RNN. The idea behind \textit{Sigmoid} Weighting is to explore a new type of hard attention, specifically in the case of ReNet, and the idea behind Directional Weight Sharing is to save parameters.
The resulting RNN cells are different from standard RNN cells, especially deeper and more sequential. 
We are able to outperform ReNet with standard RNN cells on the data sets CIFAR-10 and SVHN \cite{cifar10, netzer2011reading}, as detailed in \autoref{sec:results}.
Summarizing, our contributions are the following ones:
\begin{itemize}
\item We evaluated the DARTS approach in the ReNet model with new conditions not considered in the original paper. Thus, it is not searching RNN cells for sequential data like text, but finding an RNN cell for the visual domain.

\item We compared the novel cells with GRU and LSTM in the ReNet architecture in two small-scale data sets. The new cells reached better results on CIFAR-10 and SVHN. We want to point out that we did the cell search on the CIFAR-10 data set but evaluated on CIFAR-10 and SVHN data sets.
\item We added two more alternatives to the search space: \textit{Sigmoid} Weighting and Directional Weight Sharing.  Directional Weight Sharing achieved better results with fewer parameters on CIFAR-10 whereas the best performance for SVHN was achieved without a variant.
\end{itemize}

\section{Background}
This work relies on two crucial building blocks.
The first is ReNet, which processes images or feature maps as a sequence. 
Hence, RNN cells can be applied to images or any feature maps.
The second is DARTS.
It finds cells, which are smaller-sized structures that are usable in a Neural Network, using gradient descent.
Comparing to other NAS approaches (e.g., evolutionary methods or reinforcement learning), DARTS is fast to compute \cite{pham2018efficient, zoph2016neural, baker2016designing}. 
We combine both approaches to explore which RNN cells are suitable for image classification tasks.

\subsection{ReNet}
\label{sec:ReNet}
\begin{figure}[thb!]
    \centering
    \includegraphics[width=.7\textwidth]{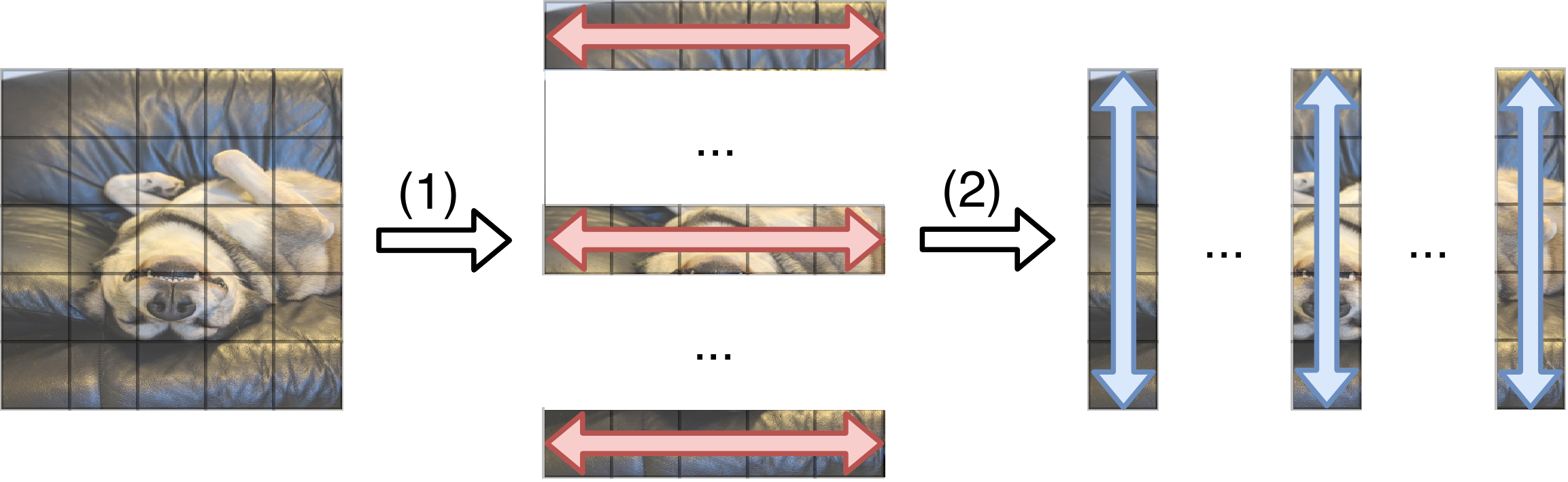}
    \caption{Example of a ReNet layer. Initially, an image is split into several patches (i.e., 2x2x3). Then, two bidirectional RNNs sweep over the image in the horizontal direction (1) and then in the vertical direction (2).}
    \label{renet_layer_fig}
\end{figure}

The ReNet layer is an RNN based alternative to convolution and pooling transformations\cite{visin2015renet}.
The basic idea is to divide a given image into flattened patches and process the patches as a sequence to bidirectional RNNs, see Fig.~\ref{renet_layer_fig}.

Let \(I \in \mathbb{R}^{ w \times h \times c}\) be an input, where $h$, $w$, and $c$ are the width, height and number of channels, respectively.
A ReNet layer with a window size of \(\left( w_p, h_p \right)\) creates a set of non-overlapping and flattened patches of $I$.
Let $\mathbf{p}_{i,j}$ be the $\left( i, j\right)$-th patch of $I$. Let $F$ and $B$ denote the two directions of a bidirectional RNN, forward and backward.
A bidirectional $RNN_H = \{RNN_H^F,  RNN_H^B\}$ reads the sequence of flattened patches in the horizontal direction of the image.
The initial hidden states are initially a zero vector $\mathbf{0}$.
It creates a feature map 
\begin{equation}
\mathbf{H} = \begin{bmatrix} 
               \mathbf{h}^F \\
               \mathbf{h}^B
               \end{bmatrix}
\end{equation} 
\noindent
with the elements
\begin{equation}
\begin{split}
\mathbf{h}^F_{i,j} &= RNN_H^F \left(\mathbf{p}_{i,j}, \mathbf{h}^F_{i,j-1}\right) \text{ and } \\ \mathbf{h}^B_{i,j} &= RNN_H^B \left(\mathbf{p}_{i,j}, \mathbf{h}^B_{i,j+1}\right) \\ 
 \label{eq:renet}
 \end{split}
\end{equation} 
with $i \in \{0,...,h\} \text{ and }  j \in \{0,...,w\}$.
Afterward, a second bidirectional $RNN_V = \{RNN_V^F,  RNN_V^B\}$ processes the feature map $\mathbf{H}$ in a similar manner but in the vertical direction.
The feature map $\mathbf{V}$ is the overall output of the ReNet layer.

Note that ReNet is agnostic to RNN cell definition (LSTM, GRU or any new development). 
Hence, it is suitable to use RNN cells derived by DARTS explained in the next section.

\subsection{DARTS}
\label{sec:DARTS}
DARTS is a cell-based NAS approach \cite{liu2018darts}. It is finding a cell design that can be used as components for a network architecture \cite{pham2018efficient}. Like Residual Blocks in ResNet \cite{he2016deep}, a cell is a small graph representing a structure that is usable within a Neural Network.
DARTS derives the cell with gradient descent in the Architecture Search Space.

We will explain two essential components of DARTS for developing a new RNN cell for ReNet: \textit{cell definition} for RNNs and \textit{cell search} based on the cell definition.
\textit{Cell definition} describes the sequence of operations within the RNN cell and the learnable elements (i.e., activation functions and connections to other internal components).
\textit{Cell search} performs gradient descent to explore the Architecture Search Space based on the \textit{cell definition}.
 
\subsubsection{Cell Definition}
The next definitions follow the standard DARTS approach from the original paper for RNN cells. A \textit{cell} is an acyclic, directed graph $G = \left(V, E\right)$ where each vertex $i \in V$ has exactly one ingoing edge $(j, i) \in E$ to a predecessor vertex $j \in V$.
The edge is associated with an activation function $f_i$. We will denote the input of a vertex $i$ with $\mathbf{\widetilde{x}}_{i, t}$.
Vertex $i$ receives the output of its predecessor as part of its input $\mathbf{\widetilde{x}}_{i, t}$.
One vertex in this graph is the input vertex, which receives an input $\mathbf{x}_{t}$ independent from the \textit{cell} itself (e.g., a flattened patch of an image).
Thus, the input $\mathbf{\widetilde{x}}_{i, t}$ of $i$ is defined as
\begin{equation}
\mathbf{\widetilde{x}}_{i, t} = 
    \begin{cases}
    \left[\begin{array}{c}
    \mathbf{x}_{t}\\
    \mathbf{h}_{t-1}\\
    \end{array} \right] & \text{ for } i=0 \\
    \enspace \mathbf{h}_{j, t-1} & \text{ for } i>0
    \end{cases}
\end{equation}
\noindent
where $j$ is denoting the predecessor vertex of $i$. The initial hidden states $\mathbf{h}_{i, 0}$ are $\mathbf{0}$. Let $\mathcal{O}$ be a set of candidate activation functions, namely \textit{Sigmoid}, \textit{Tanh}, \textit{ReLU}, and \textit{Identity}. These are the common choices of activation functions in the NAS field. It calculates two vectors: update vector $\mathbf{c}_{i, t}$ and candidate values vector $\mathbf{\widetilde{h}}_{i, t}$:
\begin{equation}
\label{c_eq}
\left(
\begin{array}{c}
\mathbf{c}_{i, t}\\
\mathbf{\widetilde{h}}_{i, t}\\
\end{array}
\right)
= 
\left(
\begin{array}{c}
\sigma\\
f_i\\
\end{array}
\right) \
\mathbf{W}_i^T \
\mathbf{\widetilde{x}}_{i, t},
\end{equation}
\noindent
with $\mathbf{W}_i$ as a learnable weight matrix and $f_i \in \mathcal{O}$, associated with an edge $e \in E$. The function and the predecessor vertex is found by DARTS during the \textit{cell} search (described below in Section: \emph{Cell Search}). An exception is given by the input vertex with $f_0 = Tanh$, also given by DARTS. The output of a vertex $i$ is 
\begin{equation}
\label{h_eq}
\mathbf{h}_{i, t} = \left( 1-\mathbf{c}_{i, t} \right) \cdot \mathbf{h}_{i, t-1} + \mathbf{c}_{i, t} \cdot \mathbf{\widetilde{h}}_{i, t}
\end{equation}
\noindent
The addable value range for $\mathbf{h}_{i, t}$ is given by $f_i$ since $\mathbf{c}_{i, t} \in \left[ 0, 1 \right]$ determines the degree in which the old hidden state is updated. Thus, the value range addable to the hidden state is depending on $\mathbf{\widetilde{h}}_{i, t}$ that itself depends on $f_i$. The addable value range becomes important in the analysis part of this paper. The overall output $\mathbf{h}_t$ of the \textit{cell} is given by 
\begin{equation}
\label{out_eq}
\mathbf{h}_t = \mathbb{E} \left[ \mathbf{h}_{i, t} \middle| i \in V \text{ and } i>0 \right].
\end{equation}

\noindent
In order to apply the \textit{cell} as a fully working RNN cell, we have to define for each vertex $i > 0$ the activation function $f_i$ and its predecessor vertex $j$. This is done automatically by DARTS during the \textit{cell} search described next.

\subsubsection{Cell Search}
\label{subsub:cellsearch}
The cell definition so far requires discrete choices of activation functions and predecessor vertices. 
DARTS initializes the graph (the cell) with all possible edges. Thus, for each vertex $i > 0$ exists one edge for each possible predecessor and activation function. We modify the cell definition from the section before by combining the outputs of all possible predecessors and all activation functions as weighted sums. This approach is called \textit{continuous relaxation} by the authors. As a result, one can use gradient descent to determine the weights and with that the most beneficial predecessor and activation function for each vertex.

It comes with heavy computational costs because of all the connections but in comparison to other NAS approaches, this approach is reliably faster w.r.t. convergence \cite{wistuba2019survey}. The weighting of all possible paths are called architecture parameters and they are trained with standard optimizers like SGD. The following is the formal description of the idea. Let $i>0$ be a vertex in the graph. In the following, 
\begin{equation}
\begin{split}
& \varphi_{j, t} \left( g,  \mathbf{x} \right) = \mathbf{h}_{j, t} \text{ with } g \in \mathcal{O}
\\
\text{s.t. } & f_j = g \text{ and } \mathbf{\widetilde{x}}_{j, t} = \mathbf{x}
\end{split}
\end{equation}
\noindent
is defined as the output $\mathbf{h}_{j, t}$ of vertex $j$ under the condition that the activation function and input is given by $g$ and $\mathbf{x}$, respectively. 
Instead of a single activation function $f_i$ of the hidden state of predecessor $j$, DARTS is considering all activations with $\tilde{f}^{(j, i)}$.
The function $\tilde{f}^{(j, i)}$ is a $\mathop{softmax}$ weighted combination of all possible activation functions instead of a discrete choice. Thus, we can use it to compute gradients over all paths. More formally, 
\begin{equation}
\tilde{f}^{(j, i)} \left( \mathbf{x} \right) = \sum_{f \in \mathcal{O}} \left[ \frac{exp \left( \alpha^{(j, i)}_f \right) }{\sum_{g \in \mathcal{O}}exp \left( \alpha^{(j, i)}_g \right)} \right] \varphi_{i, t} \left( f,  \mathbf{x} \right).
\end{equation}
The variable $\alpha^{(j, i)}_f$ is a learnable parameter that represents the weight of an edge between $i$ and $j$ that is associated with the activation function $f \in \mathcal{O}$. The discrete activation function between $i$ and any $j$ after cell search is given by the most likely activation function 
\begin{equation}
f^{(j, i)} = \argmax_{f \in \mathcal{O}} \alpha^{(j, i)}_f.
\end{equation}
\noindent
The unique predecessor vertex for the final architecture is chosen by the highest probability value of the most likely activation functions among all vertices beforehand. Thus,
\begin{equation}
\begin{split}
& \mathbf{\widetilde{x}}_{i, t} = \mathbf{h}_{j, t} \text{ and } f_i = f^{(j, i)}\\
\text{s.t. } & j = \argmax_{j < i} \left[ \max_{f \in \mathcal{O}} \alpha^{(j, i)}_f \right].
\end{split}
\end{equation}
As a result, we find a suitable activation function and predecessor for each vertex. Thus, we have a fully working RNN cell that can be used. The objective for the search is given by the task, in our case minimizing the classification loss.

\section{Methodology}

In this section, we present the methods to find new RNN cells distinguishable from standard RNN formulations like LSTM or GRU. We present the network used to find these cells. Moreover, we explain the variants we explored. 
Additionally, we describe the training process and how we used the data sets for cell search and cell evaluation. 

\subsection{Cell Search and Network Architecture}

We find new RNN cells using DARTS and ReNet, which translates an image into a sequence to apply RNN cells on images. 
Like mentioned in Section~\ref{sec:ReNet}, an arbitrary RNN cell can be used within the ReNet layer.
As a consequence, it is feasible to use DARTS to find new RNN cells specifically for the image-based domain, which has not been considered by the original DARTS paper. They have derived RNN cells for sequential data like text. 

\begin{figure}[thb!]
    \centering
    \includegraphics[width=.85\textwidth]{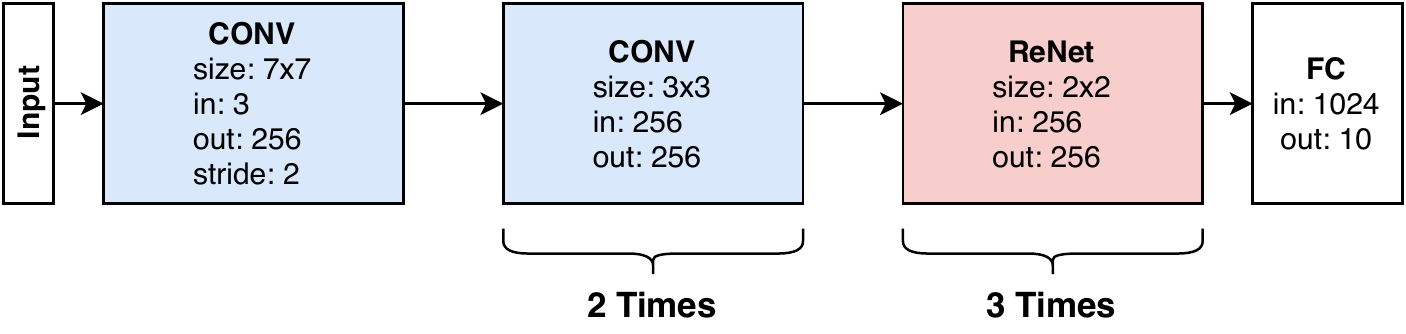}
    \caption{The network architecture used for the experiments. It consists of three convolution layers in the beginning and three ReNet layers with window size of 2 after that. The last fully connected layer is used to map the features to the number of classes. We want to point out that the ReNet layer can be defined based on the selected RNN cell (i.e., LSTM, GRU or a cell derived by DARTS).}
    \label{net_fig}
\end{figure}

However, a cell found by DARTS and used by ReNet is only a component of a network for image classification. Since we use a cell-based NAS approach, the common way is to stack ReNet layers with the found RNN cells multiple times.
Fig.~\ref{net_fig} shows the resulting network. 
It begins with three convolutional layers, followed by three ReNet layers (same number of layers as in the original paper of ReNet) with the RNN cells of DARTS.
In the end, we use a single and fully connected layer to map the feature dimension to the number of classes. This architecture is fxied for all experiments. 

The motivation behind the three convolution layers is that the original ReNet paper uses ConvZCA to whiten the input images \cite{visin2015renet, krizhevsky2009learning}.
We avoid this by using non-linear transformations of the input, realized by the convolution layers.

\subsection{Variants}

Besides ``Vanilla ReNet'' cell, we also examined two different variants.
The first variant is a \emph{Sigmoid Weighting} of the input.
We calculated a learnable weighting factor for each patch in the input sequence and took the \textit{Sigmoid} of this value to let the network choose how vital the patch is. Thus, each patch is multiplied with a value between zero and one.
It is a type of soft attention. 

The second variant combines the weights of the bidirectional RNNs in a ReNet layer similar to ``Directional Weight Sharing'' proposed in ContextVP \cite{byeon2018contextvp}. 
More formally, $RNN_V^F = RNN_V^B$ and $RNN_H^F = RNN_H^B$ for the same input, which reduces the number of parameters. 
We want to point out that even though the weights are the same, the direction still has an impact because the hidden states evolve differently during the time steps.

\subsection{Training}

All our experiments used the two data sets CIFAR-10 and SVHN, which have images of size 32x32x3, and each data set contains ten classes \cite{cifar10, netzer2011reading}. 
The data sets were normalized to have zero mean and unit variance. 
We used horizontal flipping (with a probability of 50\%) and random cropping of the original image with zero padding (size of four) and Cutout \cite{devries2017improved}.

Cell search and evaluation are two different phases. 
Given the sets $\mathit{Train}$, $\mathit{Validation}$ and $\mathit{Test}$ of CIFAR-10. 
For the cell search, we divided the $\mathit{Train}$ set s.t.\ $\mathit{Train} = \mathit{train}_{cs} \cup \mathit{val}_{cs}$, where $cs$ stands for cell search. 
We used $\mathit{val}_{cs}$ to determine the end of cell search by early stopping. 
During the cell search, we used two different Adam optimizers.
One optimizer is for the network in Figure~\ref{net_fig} (i.e., convolution and fully connected layers) and for the RNN cell parameters mentioned in the cell definition (i.e., $\mathbf{W}_i \, \forall i \in V$ in Eq.~\ref{c_eq}).
The other optimizer is for the architecture parameters in cell search step (i.e., $\alpha^{(j, i)}_f \, \forall i, j \in V$). 
The optimizing steps happen within the same batch. Thus, each optimizer works equally often.
We repeated this procedure multiple times (ca. 10 per variant) since different weight initialization can lead to different cell designs.

After training, we derived the cells as described at the end of Section~\ref{sec:DARTS}. Next, we re-initialized the weights of the network (i.e., $\mathbf{W}_i^T \, \forall i \in V$ and the weights of the convolution and fully connected layers) and trained it again on the complete $\mathit{Train}$ set of CIFAR-10. 
We used the $\mathit{Validation}$ and $\mathit{Test}$ set of CIFAR-10 to determine the end of training and the final result of the found cell (cell evaluation). 
Likewise, we also evaluated the cell on SVHN. 
No cell search was applied here — also, no transfer learning of the weights. We used only the cell designs found during cell search on CIFAR-10.
Therefore, the performance on SVHN does not explicitly benefit from the cell search.

\section{Results and Analysis}\label{sec:results}
This section presents DartsReNet cells that are beneficial for image classification along with a comparison to a standard RNN cell GRU.
Like mentioned before, there are two phases: The cell search and the cell evaluation.
Thus, this section is also divided into two sections to discuss the results of each phase. 

\subsection{Cell Search}
Here we present the RNN cells found during the cell search on CIFAR-10 using DARTS. We will discuss each variant separately.

\begin{figure}
    \centering
    \includegraphics[width=1\textwidth]{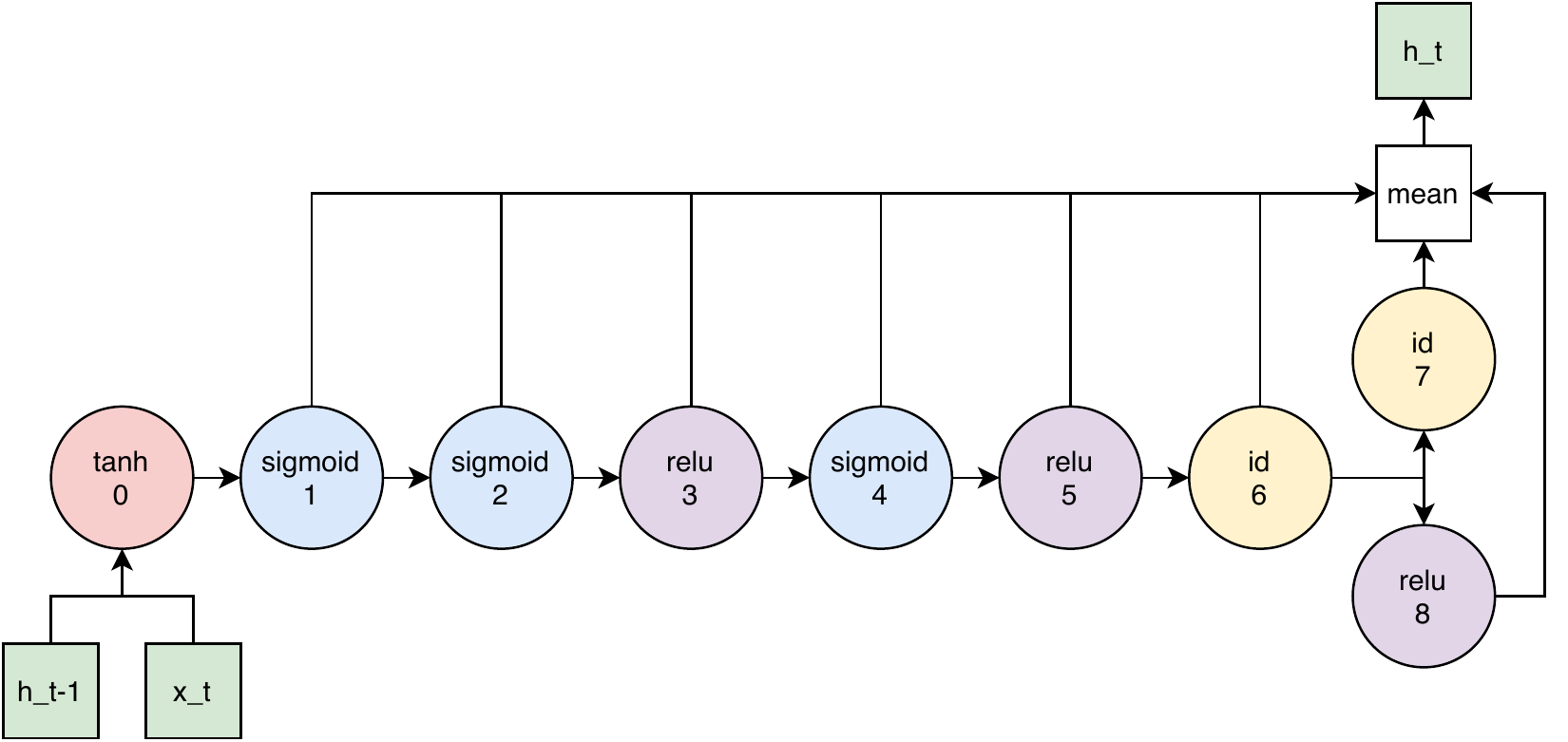}
    \caption{Derived cell for the Vanilla ReNet. It uses two \textit{Sigmoids} in the beginning, then alters between \textit{ReLU} and \textit{Sigmoid} until vertex 5. An \textit{Identity} function follows and its output is processed through another \textit{Identity} function and a \textit{ReLU} in parallel.}
    \label{cs_simple_fig}
\end{figure}

\subsubsection{Vanilla ReNet}

The derived cell is mostly sequential and deep, see Fig.~\ref{cs_simple_fig}.
The novel cell uses several \textit{Sigmoid} activation functions, two \textit{Identity} projections and \textit{ReLU}.
Surprisingly, it does not have any \textit{Tanh} activation function as the original LSTM or GRU (except the fixed input vertex, set by DARTS), which would zero center the data. 

The \textit{Sigmoid} value can be inhibiting for the mean calculation.
Consider Eq.~\ref{c_eq}, it holds that $\mathbf{c}_{i, t} \in \left[ 0, 1 \right]$ and $\mathbf{h}_{i, 0} = \mathbf{0} \, \forall i \in V$.
As a consequence of Eq.~\ref{h_eq}, the addable value range of $\mathbf{h}_{i, t}$ depends only on the value range of $\mathbf{\widetilde{h}}_{i, t}$. Hence, if $\mathbf{\widetilde{h}}_{i, t}$ is computed with $f_i = \sigma \Rightarrow \left( \mathbf{c}_{i, t} \cdot \mathbf{\widetilde{h}}_{i, t} \right) \in \left[ 0, 1 \right]$ since $\mathbf{\widetilde{h}}_{i, t} \in \left[ 0, 1 \right]$.
Therefore, \textit{Sigmoid} can add zero values to the mean calculation for $\mathbf{h}_t$, shrinking down the output, see Eq.~\ref{out_eq}.
The cell also has the possibility to add negative values to the mean value because of the \textit{Identity} functions. This is similar to the argumentation before: If $f_i = Identity$, then the value range of $\mathbf{\widetilde{h}}_{i, t}$ is in $\left[- \infty, \infty \right]$. Thus, negative values can be added to $\mathbf{h}_{i, t}$, see Eq.~\ref{h_eq}.

\begin{figure}[h!]
    \centering
    \includegraphics[width=1\textwidth]{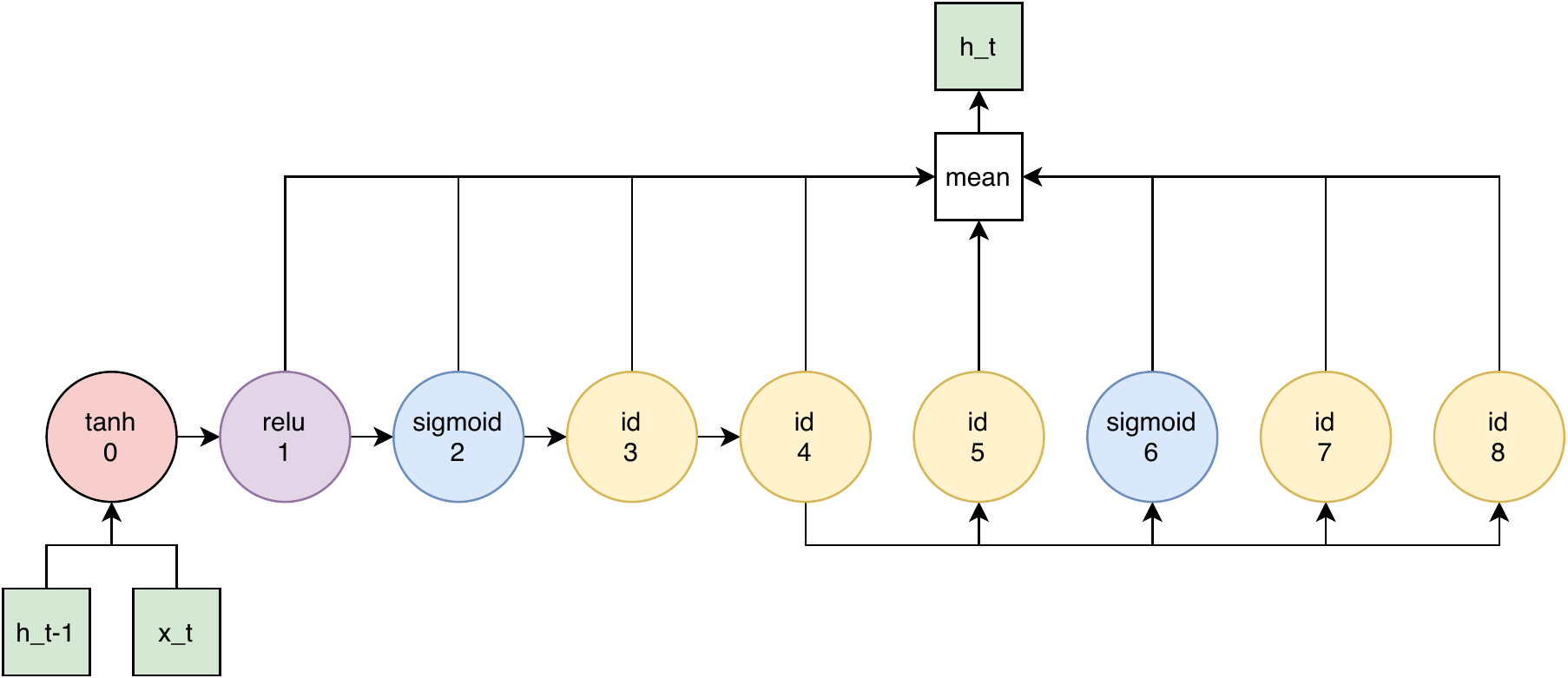}
    \caption{Derived cell for the ReNet with Sigmoid Weighting. It starts with \textit{ReLU}, \textit{Sigmoid} and \textit{Identity}. After that, it uses another \textit{Identity} and processes its output to all successor vertices, which are three \textit{Identity} and one \textit{Sigmoid} functions.}
    \label{cs_sig_fig}
\end{figure}

\subsubsection{Sigmoid Weighting}
The cell design result with $Sigmoid$ weighting is the most interesting among all three.
Fig.~\ref{cs_sig_fig} shows the found topology.
It uses no single \textit{Tanh} and surprisingly many \textit{Identity} functions.
This setup adds a lot of linear activations within the vertices for the hidden state calculation.

In contrast to the cell design in ``Vanilla ReNet'', this cell can produce more negative values through the \textit{Identities}, see Eq.~\ref{h_eq} and \ref{out_eq}.
Also, only a single \textit{ReLU} activation function is used, right in the beginning.
Additionally, this cell design uses many ``wide'' connections, starting at vertex $4$.
We mean with ``wide'' connections that more than one successor vertex use the vertex's output.
Hence, this cell is not as deep as the cell for Vanilla ReNet, supporting the assumption that the depth might be too high.

\begin{figure}[h!]
    \centering
    \includegraphics[width=1\textwidth]{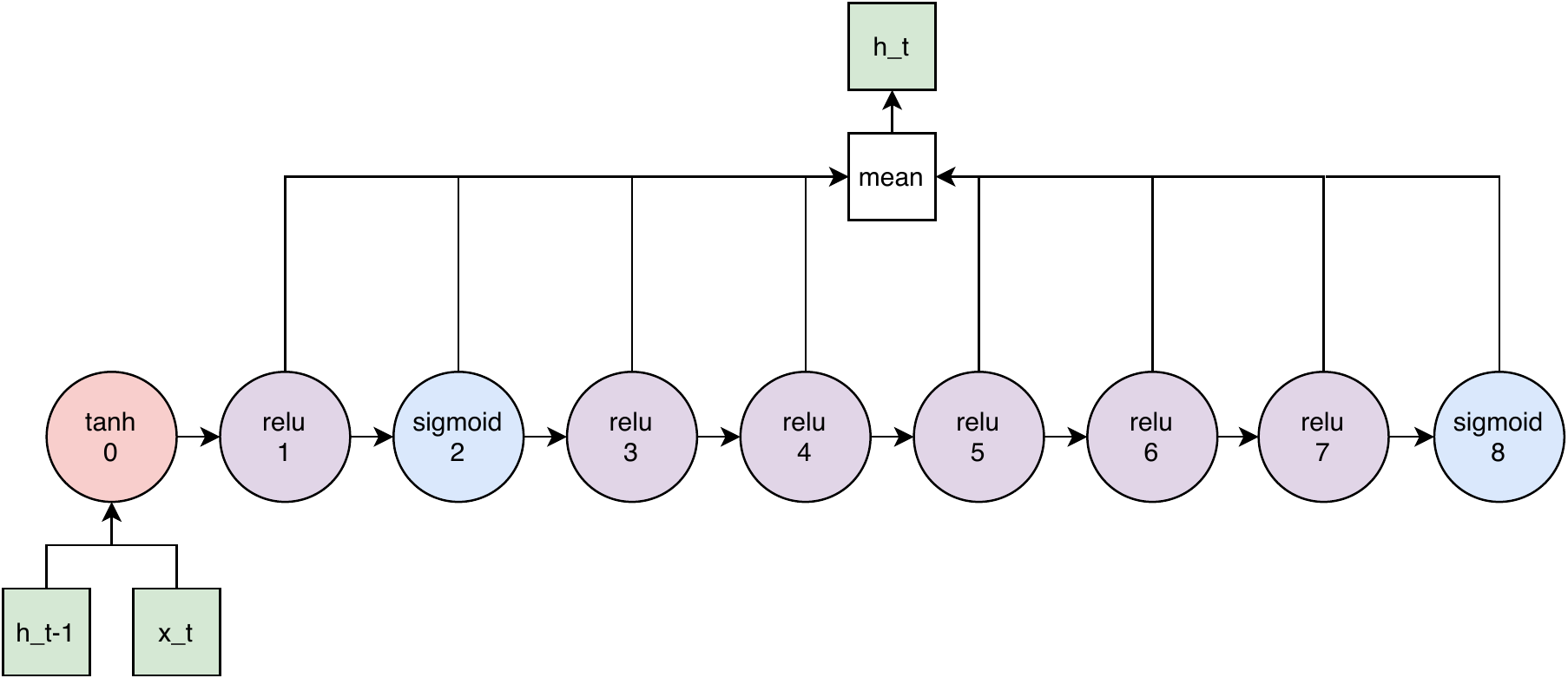}
    \caption{Derived cell for the ReNet with Directional Weight Sharing. It uses a \textit{ReLU} function in the beginning, followed by a \textit{Sigmoid}. Then, five \textit{ReLU} are used. The last vertex uses a \textit{Sigmoid}. They are applied in one sequence without sharing the output to more than one successor vertex.}
    \label{cs_para_fig}
\end{figure}

\subsubsection{Directional Weight Sharing}
The ReNet layer with Directional Weight Sharing has the same outcome as the Vanilla ReNet: It does not use \textit{Tanh}, and it is sequential.
For both variants, the tendency is to go deeper instead of wider, like for CNNs in the recent development for image classification w.r.t. architecture \cite{szegedy2015going, he2016deep}.
The cell is visualized in Fig.~\ref{cs_para_fig}. 

An exciting aspect of this design is the lack of any single \textit{Identity} function.
In conclusion, the cell has no possibility to add negative values to the mean.
Because of only two \textit{Sigmoid}, the inhibitor effect is also lower than for the Vanilla ReNet cell.

\subsection{Cell Evaluation}

The results of the cell evaluation are listed in Table~\ref{table_example} for both data sets. 
For Comparison, we also listed the results of ReNet with the standard RNN cells GRU and LSTM.
In contrast to the original ReNet paper, we have used a different setup for the ReNet with the standard RNN cells to make a fair comparison feasible.
First of all, the original paper has used no convolution layers in the beginning.
Besides, the original paper has used the whitening technique ConvZCA for CIFAR-10, which has a considerable influence on the performance. According to the channel sizes, we used a fixed channel size of 256 for both data sets instead of 320 for CIFAR-10.
Also, the original ReNet uses an additional fully connected layer for SVHN. Additionally, a fully connected layer of 4096 dimensions was used instead of 1024, which is the case for our architecture.

According to the new cells, all variants show better accuracy for all data sets. For CIFAR-10, the variant Directional Weight Sharing dominates. For SVHN, the Vanilla variant has the best accuracy. Because of the depth of the new cells, all variants have a higher parameter size w.r.t. GRU. An exception is the Directional Weight Sharing, which uses 0.8M fewer parameters than the baseline with GRU. Since the variants work like regularization to the training, one can see ascending accuracies among the variants for CIFAR-10. The regularization is also present for SVHN, where our Vanilla ReNet has the best performance. The complexity is high enough for Vanilla ReNet and additional regularization hurts the performance.  

\begin{table}[!t]
\renewcommand{\arraystretch}{1.3}
\caption{Results of the experiments. Cell Search was applied to CIFAR-10 and the derived cells were evaluated on CIFAR-10 and SVHN. The accuracy of the ReNet with the derived cell, its variants and  ReNet with the standard cells GRU and LSTM. Also listed is the parameter size of each model.}
\label{table_example}
\centering
\begin{tabular}{llrr}
\hline
\bfseries Data Set & \bfseries Model & \bfseries Params [M] & \bfseries Accuracy [\%]\\
\hline
CIFAR-10 & Vanilla - DartsReNet & 6.8 & 90.26\\
 & Sigmoid Weighting - DartsReNet & 6.8 & 90.56\\
 & \textbf{Directional Weight Sharing - DartsReNet} & \textbf{4.0} & \textbf{91.00}\\

\cline{2-4}
 & Baseline ReNet with GRU  & 4.8 & 75.30\\
 & Baseline ReNet with LSTM  & 6.0 & 74.94\\
\hline
SVHN & \textbf{Vanilla - DartsReNet} & \textbf{6.8} & \textbf{97.43}\\
 & Sigmoid Weighting - DartsReNet & 6.8 & 97.04\\
 & Directional Weight Sharing - DartsReNet & 4.0 & 96.91\\
\cline{2-4}
 & Baseline ReNet with GRU  & 4.8 & 95.16 \\
  & Baseline ReNet with LSTM  & 6.0 & 94.10
\end{tabular}
\end{table}

\section{Conclusion and Future Work}

We wanted to examine alternative RNN cell designs for image classification other than standard formulations like LSTM or GRU.
We achieved this by combining DARTS with ReNet.
Moreover, we examined two different variants of RNN cells using this approach.
Also, we compared the cells with a ReNet architecture with GRU and LSTM on CIFAR-10 and SVHN.
We achieved better results by a large margin and the variant Directional Weight Sharing achieved this with 0.8M fewer parameters than GRU.
Additionally, the RNN cell, found on CIFAR-10, is also evaluated on SVNH data set, where we were able to outperform the standard cells by at least 1\%.

For the future, we are interested in evaluating our approach on ImageNet \cite{imagenet_cvpr09}.
Also, this work only used NAS approaches to derive cells with a fixed network.
However, it is possible to use it for the overall network design. 
Some other variants are also interesting for future investigation, like different ordering within a sequence or dropout parts of the image as learnable components.

\section*{Acknowledgement}
This work was supported by the BMBF project DeFuseNN (Grant 01IW17002) and the NVIDIA AI Lab (NVAIL) program.

\bibliographystyle{splncs04}
\bibliography{icann2019}
\end{document}